# Care Robots with Sexual Assistance Functions

Oliver Bendel

School of Business FHNW, Bahnhofstrasse 6, CH-5210 Windisch
oliver.bendel@fhnw.ch

## Abstract

Residents in retirement and nursing homes have sexual needs just like other people. However, the semi-public situation makes it difficult for them to satisfy these existential concerns. In addition, they may not be able to meet a suitable partner or find it difficult to have a relationship for mental or physical reasons. People who live or are cared for at home can also be affected by this problem. Perhaps they can host someone more easily and discreetly than the residents of a health facility, but some elderly and disabled people may be restricted in some ways. This article examines the opportunities and risks that arise with regard to care robots with sexual assistance functions. First of all, it deals with sexual well-being. Then it presents robotic systems ranging from sex robots to care robots. Finally, the focus is on care robots, with the author exploring technical and design issues. A brief ethical discussion completes the article. The result is that care robots with sexual assistance functions could be an enrichment of the everyday life of people in need of care, but that we also have to consider some technical, design and moral aspects.

## Introduction

Residents in retirement and nursing homes have sexual needs just like other people. However, the semi-public situation makes it difficult for them to satisfy these existential concerns. They may also find it challenging to find a suitable partner, or it may be mentally or physically difficult for them to have sexual intercourse or a relationship. People who live or are cared for at home (Giuliani et al. 2005) may also be affected by this problem. Perhaps they can meet someone more easily and discreetly than the residents of a health facility, but some elderly and disabled people may be restricted in some ways.

This article examines the opportunities and risks that arise with regard to care robots with sexual assistance functions. First, it deals with sexual well-being. Second, it presents robotic systems ranging from sex robots to care robots. Third, the focus is on care robots, with the author investigating technical and design issues. A short ethical discussion and a summary with an outlook round off the article.

## Sexual Well-being

The Geneva-based World Health Organization (WHO) believes that sexual health is more than the absence of sexually transmitted diseases, sexual dysfunction and sexual violence. Rather, it includes sexual well-being (WHO 2006). According to Döring 2018, sexual well-being contributes to overall mental and physical health through various mechanisms (e.g. testosterone production, stress reduction, pain reduction, partnership commitment).

We can even regard sexual well-being as a human right. The aforementioned WHO, the United Nations (UN), the International Planned Parenthood Federation (IPPF) and the World Association for Sexual Health (WAS) have developed and published declarations of sexual and reproductive human rights (Döring 2018). Of course, it is not a basic right to have sex anytime, anywhere, or with anyone. But it is about basic access to it.

There is no reason to exclude the elderly and disabled or those in need of care from sexual well-being, and if you recognize that this is a human right, it goes without saying that all people who want to be sexually active should be considered. This, of course, takes place within the current social and legal framework, so that, for example, the sexual wishes of minors must also be rejected.

There has been little research into what people in need of care, the elderly and the disabled really want. Sexuality in the care sector seems to be a taboo topic, for scientists as well as affected persons, for individuals as well as for organizations. Rosalind Mary Owens, known as Tuppy Owens, is the founder of Outsiders, a private club providing peer support and dating opportunities for physically

---

and socially disabled people, and set up the Sexual Health and Disability Alliance (SHADA) for health and social care professionals. According to her, people with disabilities – the focus of her book "Supporting Disabled People with their Sexual Lives: A Clear Guide for Health and Social Care Professionals" – have four central concerns regarding their sexual life (Owens 2015):

A. "Feel sexually free."
B. "Become educated about sex and their own sexuality."
C. "Enjoy sexual activities."
D. "Support in a crisis."

The main focus here will be on concern C, i.e. the actual sexual activities. But the author of this article also dedicates a section to concern B. According to Owens (2015), the following actions are decisive for solo and partner sexual activities from the point of view of people with disabilities:

1. "Being hugged (if and when appropriate)"
2. "Sharing their bed"
3. "Using pornography"
4. "Using sex toys"
5. "Partnered sex"
6. "Using sexual services for reasons other than education"
7. "Finding financial support for sex products and services"

According to Döring (2018), the needs of elderly and old people are described similarly. In particular, the idea that older people live largely asexually is incorrect. Even if some of them have fewer opportunities, their desires and impulses remain largely or completely preserved.

If one now takes concern C seriously and considers several of the seven points listed, also and especially against the background of the outlined situation in which persons in need of care, the elderly and the sick find themselves, the inclusion of sex robots and care robots with sexual assistance functions presents itself as an option. In the following, the author distinguishes relevant types and identifies concrete products in order to obtain an overview of what exists in reality.

## Relevant Robot Types

### Sex Robots and Love Dolls

Love dolls in the stricter sense are common, sex robots are not. Still, there is much more literature about sex robots (Levy 2008; Cheok et al. 2017; Cheok and Levy 2018; Zhou and Fischer 2019). They seem to stimulate our imagination more. There are only a few sex robots on the market, and their mobility is limited. Like love dolls you have to position them in a suitable place.

TrueCompanion's often mentioned Roxxxy is a sex robot that, according to the company, can listen, talk and respond to touch (Bendel 2015). You can choose between different personalities, from "Wild Wendy" to "Frigid Farrah", and between different sexual orientations. The male counterpart is Rocky. You could allegedly order both from the website (www.truecompanion.com), which advertised with the words "world's first sex robot", also using the term "sex robot doll". There have been speculations that the two robots do not exist, at least not as a product, and the website is now offline.

Shenzhen All Intelligent Robot Technology (www.ai-aitech.com), Doll Sweet (www.dsdolleurope.com), Abyss Creations (www.realdoll.com) and the affiliated well-known "factory" Realbotix (realbotix.com) launch love dolls, also known as dolls or sex dolls (Bendel 2019a, 2019b). Early on, these figures attracted attention with their detailed and knowledgeable, albeit idealizing, design. They have silicone skin or skin made of thermo-plastic elastomers. Gel can be stored underneath so that the body parts feel real, like the elastic flesh of young women and men. Love dolls sometimes have synthetic voices or simple motor skills such as turning the head and moving the eyes. Shenzhen All Intelligent Robot Technology stresses that artificial intelligence (AI) is in its products, and there may be a display in the back of the head.

All in all, the boundaries between love dolls and sex robots are fluid. The love dolls currently found in the brothels of Germany, Spain, Russia and Switzerland are not sex robots. However, if you extend them with mimic and gestural abilities and with natural language abilities, the term is suddenly more appropriate. Harmony by Realbotix has left the existence of classic rubber dolls far behind. She has mimic abilities, and she speaks to the user and understands him or her within the bounds of her possibilities. The company from San Diego also stresses the fact that AI is used – Harmony herself emphasizes this in videos (Inventions World 2017). For sex robots as well as for love dolls it is true that they are usually designed human-like, and some advanced models can be seen as androids.

### Care Robots

The majority of care robots are prototypes. You find them in test environments as well as at trade fairs, conferences and exhibitions (Becker 2018; Bendel 2018a). Tests and studies also take place in hospitals and nursing homes. Early developments include the "nurse's assistant" HelpMate and the "nurse-bot" Pearl, which support nursing staff (Bekey 2012). HelpMate transports things, Pearl provides useful information and visits patients. JACO² 6 DOF by Kinova Robotics (kinova-robotics.com), short JACO, can help people with limited arm and hand functions. It consists of one arm and one hand with three fingers, which distinguishes it from conventional cooperation and collaboration robots with their two fingers; nevertheless, it is very close to these co-robots, for example with regard to the basic construction and the number of degrees of freedom.

Care-O-bot from Fraunhofer Institute for Manufacturing Engineering and Automation IPA in Stuttgart, Germany (www.care-o-bot.de) is able to grab things and bring them away and moves safely among people and through spaces. It has a gripper with which it can pick up and hand over a flower, for example. In an advertising video, it does just that, and when the other person smiles, it simulates emotional reactions on its big display in the head area. Aethon's TUG has similar goals and functions (www.aethon.com). It is able to transport medicines and materials and take the elevator; it also has natural language skills. The HOBBIT, designed as a mobile information terminal within an EU project, was intended to help senior citizens (hobbit.acin.tuwien.ac.at). It should strengthen the feeling of safety and be able to pick up objects from the ground. Cody from the Georgia Institute of Technology (College of Engineering, coe.gatech.edu) can turn bedridden patients around and wash them.

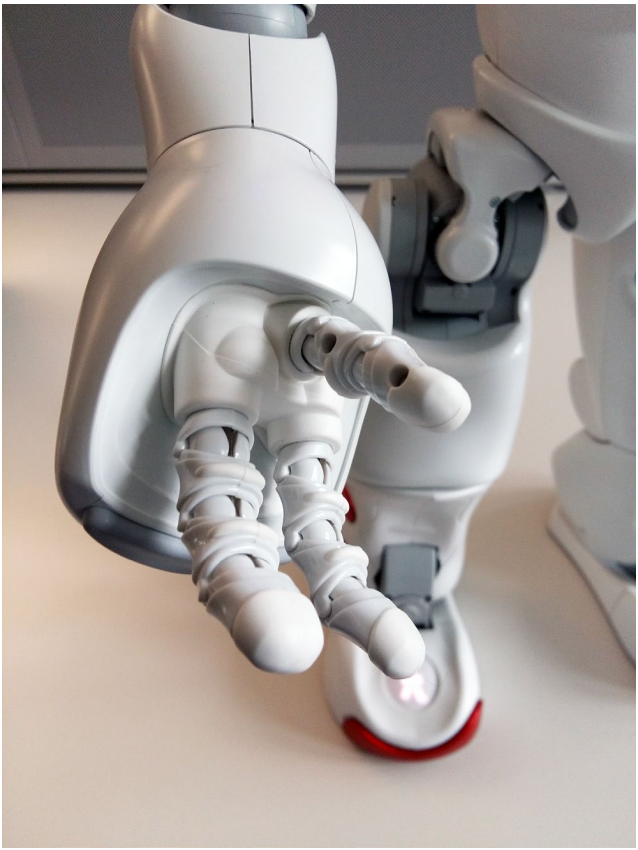

Fig. 1: The fingers of Nao

The Japanese Robear (predecessor versions RIBA and RIBA-II) from Riken (www.riken.jp) – as bearish in appearance as the name suggests – works in tandem with the caregiver and assists with the transfer from and to the bed and with the straightening in the bed. The research institution discontinued further development, although there were promising approaches and tests. TWENDY-ONE, a humanoid robot with two hands with four fingers each from the Sugano Laboratory of WASEDA University, supports patients with helping them sit up and household tasks (twendyone.com).

F&P Personal Robotics (www.fp-robotics.com) based in Switzerland has developed the service robots Lio and P-Rob. They are similar to JACO, but have usually only two fingers. Basically, they are co-robots specialized in the health sector. Their interchangeable attachments at the end of the arm allow them to be used for both therapy and care. A new product was developed in cooperation with a Chinese partner: P-Care. It has two arms, an actual body and a head that resembles that of a monkey, more precisely the Monkey King (Sūn Wùkōng in the classic Chinese novel). Lio has been repeatedly tested in old people's and nursing homes in Switzerland and Germany (Früh and Gasser 2018; Wirth 2020).

More and more often, the humanoid robots of Aldebaran (today part of SoftBank), namely the small Nao (with legs and hands with three fingers, see Fig. 1) and the larger Pepper (with castors and hands with five fingers, see Fig. 2), are found in retirement and nursing homes or in corresponding simulated environments. Anderson et al. (2019) used Nao in a machine ethics project. Pepper often functions as an information and entertainment robot. Although its arms and hands are designed down to the last detail and are almost like those of human beings, it is hardly suitable as a transport robot and for receiving and serving medicine and food – the extremities are supposed to enable convincing gestures, no more and no less. The same applies to the smaller Nao.

It is obvious that care robots are sometimes abstract, sometimes animal- or human-like. Robear and P-Care can certainly be described as intermediate forms. They resemble, as already said, a bear or a monkey, but Donald Duck is also created after the model of a duck and yet appears more like a human being (the same goes for Mickey Mouse who is more a human than a mouse). This is due to the design of body parts, such as arm and hand, behavior, and natural language skills (visually depicted in the comic, via speech bubbles, implemented via sound in the animated film). Even if a robot does not have such abilities – the mere fact that it takes on tasks that arise in nursing and support brings it closer to people in the perception of those in need of care.

# Robots with Sexual Connotations

## Sex Robots and Love Dolls

Sex robots and love dolls can certainly be used in old people's and nursing homes. However, there are several points against this:

– They are strongly sexualized, which can be experienced as inappropriate in this context.
– They partly correspond to stereotypes that are perceived differently and sometimes rejected.
– They are used in controversial environments such as brothels.
– They tend to be aimed at private buyers and casual customers and are not age- and disability-friendly.
– They are primarily aimed at men, in both female and male forms.

Presumably, the use of sex robots and love dolls would currently meet with little acceptance among those in need of care, nursing staff and relatives. A similar case can be assumed for use at home. In addition to the caregivers and relatives, friends and neighbors could also be disturbed by the presence of the sex robot. Nevertheless, under certain circumstances there might be possible solutions:

– One could place the sex robots and love dolls in separate rooms where they cannot be noticed and can be discreetly visited.
– The persons responsible would only take them out on certain occasions in order to lend them to or bring them to the users.
– Certain types could be adapted, softened or defused by appropriate clothing or body interventions.
– Manufacturers could be asked to consider women as users, for example with a greater variety of male and female figures.
– One could work towards a change in posture and make it clear that those affected have the right to use sex robots with the look they want.

Another possibility would be to equip care robots with sexual assistance functions. The author will discuss this in the following.

## Care Robots with Sexual Assistance Functions

Instead of a sex robot or a love doll, care robots with sexual assistance functions could also be used, both in retirement and nursing homes and within one's own home. They would initially refer to concern C and to three of the Owens points mentioned (2015), namely "Being hugged", "Sharing their bed" and "Partnered sex".

Embraces by robots are possible if they have two arms, such as P-Care, restricted also with one arm. However, the hugs and touches feel different to those made by humans. When one uses warmth and softness, like in the HuggieBot project, the effect improves, but is still not the same (Block and Kuchenbecker 2019). In hugs it is important that another person hugs us (hugging ourselves is totally different), and that this person is in a certain relationship to us. He or she may be strange to us, but there must be trust or desire. Whether this is the case with a robot must be assessed on a case-by-case basis. There is some evidence that two arms are not enough, that instead humanoid forms of appearance and behavior must be added.

It is also possible for robots and humans to be in the same bed. Here, too, the effects are unclear. A human being might want to experience security and well-being, and it is doubtful whether the robot can convey these. If the robot simulates breathing movements or snoring, this could help, depending on the patient's condition. But all in all, many people affected will notice that there is only a thing next to them, similar to a plush toy next to a child, only that the adolescent eventually develops a distance to his or her artificial companion and disposes of it. Nevertheless, this get-together is not completely absurd. The Somnox sleep robot (meet-somnox.com) aims to eliminate insomnia, support natural recovery and reduce stress (Samadder 2019). The kidney-shaped cushion gently breathes in and out and plays soothing sounds.

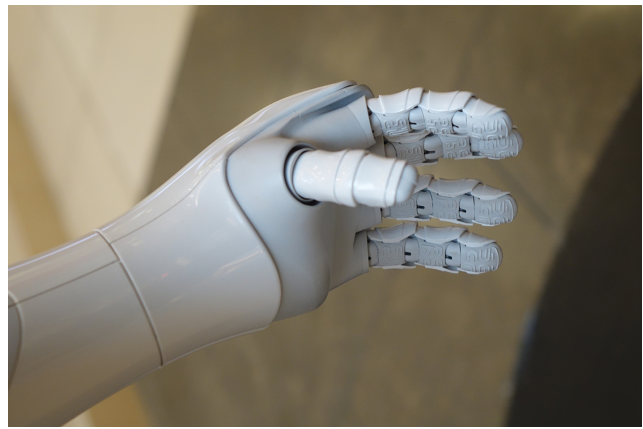

Fig. 2: The hand of Pepper

A care robot could contribute to partnered sex through direct stimulation with the help of its extremities or through openings through which it can be penetrated. The tools and openings can be more discreet than with sex robots. The fact that such use cannot be ruled out even today is shown by SoftBank's indication that one should not have sex with Pepper (Lott-Lavigna 2015). Its hands might actually be suitable for masturbation (see Fig. 2); they might just need to be a little more powerful and to use some liquids or gels. Nao's hands could also be helpful, for both sexes (see Fig. 1). The "upgrade" of the care robot would take place with hardware and software. A robot such as Lio or P-Care could receive new end pieces (particularly convincing hands or forms of sex toys) – or appropriate movement possibilities thanks to new software or newly learned actions.

If patients have movement restrictions, a care robot with a robot arm such as Care-O-bot or Twendy-One according to Döring (2018) can also be used to give the patient a sex toy, position it and hold it. Related to Owens, this would be a support at point 4 (you are still at concern C). With such robotic manipulations, the person in need of care would be more independent in living their solo sexuality and would not have to ask relatives or caregivers for help with this intimate matter (Döring 2018).

The care robot could also discreetly take over the cleaning of the sex toy, an important task for which hardly anyone feels responsible in everyday care (Döring 2018). Here, too, point 4 of Owens is addressed, and in the broadest sense, still concern C. In addition, there is an aspect of sexual health that has hardly been appreciated so far, namely the avoidance of sexual diseases. One has to consider the possibility that the sex toy is used by several people and therefore there is a risk of infection.

Of course, the care robot can also – point 3 of Owens – be used to provide and dispose of pornography (concern C). It may also be used as an extended audio book, i.e. it can read erotic and pornographic texts and make noises such as moaning and sighing. The marking language SSML is only rudimentarily suitable for this and would have to be extended. Bendel (2018b) has made proposals for this. The robot can also play along a bit, with appropriate facial expressions and gestures as well as body movements. It is difficult to say whether all this is desired by those affected and would have to be tested.

Basically, natural language skills are an important option. According to the website, the manufacturer attached great importance to communication when equipping Roxxxy. Even those who are no longer physically active can enjoy stimulating conversations or dirty talk (Bendel 2018b). AI systems could be used to get to know the user and his or her preferences better and better and to satisfy him or her verbally – and to address him or her accurately with the name (perhaps also a nickname).

Also concern B can be developed with the help of care robots with sexual assistance functions. They could thus offer sexual education, and perhaps this form would help to avoid shame one would have towards humans (Bendel 2018a). Those in need of care could ask all sorts of questions that the care robot could answer with the help of an extensive knowledge base and as a networked system, professionally flawlessly and with robotic neutrality or looseness. In the background, a caregiver or a doctor could help in individual cases without the patient having to notice this.

According to Döring (2018), to think about sexual assistance functions from the outset when designing care robots and to develop them together with the target groups requires all participants to be guided by the parameters of sexual health and sexual human rights (see also Barrett and Hinchcliff 2018). Otherwise, she adds, the issue of sexuality will remain stigmatized and marginalized. So there is a demand for education, for attention to the real needs of patients and elderly persons, and for the willingness to do the right thing to do the good thing, even if it is difficult.

## Brief Ethical Discussion

In this context – the use of sex robots and love dolls or machine sexual assistance functions for people in need of care is a novelty and in the eyes of many people a curiosity – an ethical discussion must undoubtedly be conducted. It must not be a question of moralizing or patronizing – which in principle is not the task of philosophical ethics. Rather, dialectical and discursive methods have to be used to determine whether, for example, general human rights are violated or whether certain conflicts arise (e.g. with special personal rights).

At first glance, one could argue that care robots exhibiting sexual assistance functions offend human dignity. In fact, many people will reject and even find repulsive the use of such robots, they will assume these systems are deceptive – or unnatural, even if this is a bad argument. So it is all the more important that the use is voluntary. But one could also question whether robots really promote sexual well-being – or rather destroy it. Ultimately, empirical studies must show this.

Some people will be bothered by the fact that their relatives are fobbed off with machine options in their sexual needs. Others will deny the sexual needs of their old or dependent acquaintances and relatives at all. Reservations are also likely to emerge among those in need of care themselves. Sex robots and love dolls, as well as care robots with a sexual assistance function, are capable of triggering violent disputes in nursing homes, hospitals and private households. One can try to soften or dissolve them in discourse.

Furthermore, intimacy and privacy and the related informational autonomy are important issues. Robots usually have cameras and sensors of all kinds, especially when they are mobile, and can thus, in principle, photograph and monitor patients in need of care. Personal data in this area is very sensitive and must be protected. Self-learning systems pose further problems. The AI-enhanced robot could create detailed profiles that would be interesting for nefarious groups, for example for coercion and extortion.

An alternative would be sexual assistance by human specialists. Döring (2018) lists sexual counselling, sexual support and sex work. Sexual counselling creates conditions and prerequisites for sexual activities without the professional him- or herself being actively involved in sexual activities. Sexual accompaniment in the sense of active sexual assistance enables sensual-sexual closeness with the specialist, who is often referred to as the sexual accompanist. Last but not least, the erotic industry is adapting to the growing number of older people and people in need of care – in other words, it is classic sex work with a new target audience. Of course, others will be disturbed by the fact that people – usually women – offer sexual activities here, even if they do so voluntarily.

## Summary and Outlook

The present paper examined the possibilities that arise in the field of sexual well-being with regard to care robots. It turned out that there is potential to this idea. Depending on the country and culture, sex robots and love dolls may appear inappropriate in old people's and nursing homes (or in assisted living). Care robots will spread more and more, and one could upgrade them more or less inconspicuous in hardware and software. Of course, controversies are unavoidable even then.

Whether care robots, as they currently present themselves, are really suitable for sexual assistance functions related to partner sexuality is not clear. Some like JACO, Lio and P-Rob seem too technical, others like HOBBIT too experimental, others like Robear too "animaloid". Bendel (2019b) has pointed out that some younger men in brothels with love dolls may be attracted to fantasy figures, but that Daisy Duck or Minnie Mouse are probably the limit (or Donald Duck and Mickey Mouse to cover up all the customs). This is perhaps not due to the animal part, but to the comic-like nature and the fact that they are childhood figures.

Sex robots and love dolls seem to be more likely candidates due to the far-reaching humanoid design, which makes them androids. There is no doubt that we are naturally attracted to attractive people, and robots that meet certain criteria may awaken similar feelings in us. Harmony, for example, looks desirable, at least if you prefer certain stereotypes like full lips and big breasts. However, the development of sex toys over the last ten years has been different. Whereas in the past the primary sexual characteristics were reproduced – think of dildos and artificial vaginas –, today more and more abstract or thing-like forms are preferred. It is about stimulation, e.g. with vibrations or sounds. In this respect, non-humanoid care robots could also satisfy sexual needs.

What is needed are qualitative and quantitative studies and empirical research that sheds light on the situation. A prerequisite, however, would be that care robots be equipped with sexual assistance functions and tested with persons in need of care. The future must show whether science and practice are prepared to do this.